\documentclass{article}

\PassOptionsToPackage{numbers, compress}{natbib}

\usepackage[final]{neurips_2023}

\usepackage[utf8]{inputenc} 
\usepackage[T1]{fontenc}    \usepackage{hyperref}
\usepackage{url}     
\usepackage{booktabs}
\usepackage{amsfonts}
\usepackage{nicefrac}
\usepackage{microtype} 
\usepackage{xcolor}
\usepackage{graphicx}
\usepackage{import}
\usepackage{multirow}

\newcommand{\be}{\begin{equation}}
\newcommand{\ee}{\end{equation}}
\newcommand{\ba}{\begin{align}}
\newcommand{\ea}{\end{align}}
\newcommand{\bea}{\begin{eqnarray}}
\newcommand{\eea}{\end{eqnarray}}

\usepackage{amsmath}

\newcommand{\bh}{{\mathbf{h}}}

\newcommand{\bk}{{\mathbf{k}}}

\newcommand{\bm}{{\mathbf{m}}}

\newcommand{\bq}{{\mathbf{q}}}

\newcommand{\bx}{{\mathbf{x}}}
\newcommand{\by}{{\mathbf{y}}}

\newcommand{\bH}{{\mathbf{H}}}

\newcommand{\bK}{{\mathbf{K}}}

\newcommand{\bQ}{{\mathbf{Q}}}

\newcommand{\bW}{{\mathbf{W}}}
\newcommand{\bX}{{\mathbf{X}}}

\title{Hierarchical Joint Graph Learning and Multivariate Time Series Forecasting}

\author{%
  Juhyeon Kim, Hyungeun Lee, Seungwon Yu, Ung Hwang, Wooyul Jung, Miseon Park \\
  Department of Electronic Engineering\\
  Hanyang University, Seoul, Korea \\
  \texttt{\{wngus1310,didls1228,ahddl1324,gogowinner,jung6231\}@hanyang.ac.kr}\\
  \texttt{misun9631@naver.com}
  \AND
  Kijung Yoon\\
  Department of Electronic Engineering\\ Department of Artificial Intelligence \\
  Hanyang University, Seoul, Korea \\
  \texttt{kiyoon@hanyang.ac.kr}
}

\begin{document}
\maketitle

\begin{abstract}
  Multivariate time series is prevalent in many scientific and industrial domains. Modeling multivariate signals is challenging due to their long-range temporal dependencies and intricate interactions--both direct and indirect. To confront these complexities, we introduce a method of representing multivariate signals as nodes in a graph with edges indicating interdependency between them. Specifically, we leverage graph neural networks (GNN) and attention mechanisms to efficiently learn the underlying relationships within the time series data. Moreover, we suggest employing hierarchical signal decompositions running over the graphs to capture multiple spatial dependencies. The effectiveness of our proposed model is evaluated across various real-world benchmark datasets designed for long-term forecasting tasks. The results consistently showcase the superiority of our model, achieving an average 23\% reduction in mean squared error (MSE) compared to existing models.
\end{abstract}

\section{Introduction}
\label{sec:introduction}
Multivariate time series forecasting is a primary machine learning task in both scientific research and industrial applications \citep{petropoulos2022forecasting,lam2022graphcast}. The interactions and dependencies between many time series data govern how they evolve, and these can range from simple linear correlations to complex relationships such as the traffic flows underlying intelligent transportation systems \citep{derrow2021eta,rahmani2023graph,wu2020connecting,shang2021discrete} or physical forces affecting the trajectories of objects in space \citep{battaglia2016interaction,kipf2018neural,sanchez2020learning}.

Accurately predicting future values of the time series may require understanding their true relationships, which can provide valuable insights into the system represented by the time series. Recent studies aim to jointly infer these relationships and learn to forecast in an end-to-end manner, even without prior knowledge of the underlying graph \citep{wu2020connecting,deng2021graph}. However, inferring the graph from numerous time series data has a quadratic computational complexity, making it prohibitively expensive to scale to a large number of time signals.

Another important aspect of time series forecasting is the presence of non-stationary properties, such as seasonal effects, trends, and other structures that depend on the time index \citep{cleveland1990stl}. Such properties may need to be eliminated before modeling, and a recent line of work aims to incorporate trend and seasonality decomposition into the model architecture to simplify the prediction process \citep{oreshkin2020N-BEATS,challu2023nhits}.

Therefore, it is natural to ask whether one can leverage deep neural networks to combine the strength of both worlds: 1) using a latent graph structure that aids in time series forecasting with each signal represented as a node and the interactions between them as edges, and 2) using end-to-end training to model the time series by decomposing it into multiple levels, which enables separate modeling of different patterns at each level, and then combining them to make accurate predictions. Existing works have not addressed both of these strengths together in a unified framework, and this is precisely the research question we seek to address in our current study. 

To address this, we propose the use of graph neural networks (GNN) and a self-attention mechanism that efficiently infers latent graph structures with a time complexity and memory usage of $\mathcal{O}(N\log N)$ where $N$ is the number of time series. We further incorporate hierarchical residual blocks to learn backcast and forecast outputs. These blocks operate across multiple inferred graphs, and the aggregated forecasts contribute to producing the final prediction. By implementing this approach, we have achieved a superior forecasting performance compared to baseline models, with an average enhancement of 23\%. For an overview, this paper brings the following contributions:

\begin{enumerate}
    \item We introduce a novel approach that extends hierarchical signal decomposition, merging it with concurrent hierarchical latent graphs learning. This is termed as hierarchical joint graph learning and multivariate time series forecasting (HGMTS).
    \item Our method incorporates a sparse self-attention mechanism, which we establish as a good inductive bias when learning on latent graphs and addressing long sequence time series forecasting (LSTF) challenges.
    \item Through our experimental findings, it is evident that our proposed model outperforms traditional transformer networks in multivariate time series forecasting. The design not only sets a superior standard for direct multi-step forecasting but also establishes itself as a promising spatio-temporal GNN benchmark for subsequent studies bridging latent graph learning and time series forecasting.
\end{enumerate}
    
\section{Related Work}
    Until recently, deep learning methods for time series forecasting have primarily focused on utilizing recurrent neural networks (RNN) and their variants to develop a sequence-to-sequence prediction approach \citep{yu2017long, qin2017dual, wen2017multi, salinas2020deepar}, which has shown remarkable outcomes. Despite significant progress, however, these methods are yet to achieve accurate predictions for long sequence time series forecasting due to challenges such as the accumulation of errors in many steps of unrolling, as well as vanishing gradients and memory limitations \citep{sutskever2014sequence}.
    
    Self-attention based transformer models proposed recently for LSTF tasks have revolutionized time series prediction and attained remarkable success. In contrast to traditional RNN models, transformers have exhibited superior capability in capturing long-range temporal dependencies. Still, recent advancements in this domain, as illustrated by LongFormer \citep{beltagy2020longformer}, Reformer \citep{Kitaev2020Reformer}, Informer \citep{zhou2021informer}, AutoFormer \citep{chen2021autoformer}, and ETSformer \citep{woo2022etsformer}, have predominantly zeroed in on improving the efficiency of the self-attention mechanism, particularly for handling long input and output sequences. Concurrently, there has been a rise in the development of attention-free architectures, as seen in \citet{oreshkin2020N-BEATS} and \citet{challu2023nhits}, which present a computationally efficient alternative for modeling extensive input-output relationships by using deep stacks of fully connected layers. However, such models often overlook the intricate interactions between signals in multivariate time series data, tending to process each time series independently.     
   
    Spatio-temporal graph neural networks (ST-GNNs) are a specific type of GNNs that are tailored to handle both time series data and their interactions. They have been used in a wide range of applications such as action recognition \citep{yan2018spatial,huang2020spatio} and traffic forecasting \citep{li2017diffusion,seo2018structured,zhao2019t}. These networks integrate sequential models for capturing temporal dependencies with GNNs employed to encapsulate spatial correlations among distinct nodes. However, a caveat with ST-GNNs is that they necessitate prior information regarding structural connectivity to depict the interrelations in time series data. This can be a limitation in cases where the structural information is not available.
    
    Accordingly, GNNs that include structure learning components have been developed to learn effective graph structures suitable for time series forecasting. Two such models, NRI \citep{kipf2018neural} and GTS \citep{shang2021discrete}, calculate the probability of an edge between nodes using pairwise scores, resulting in a discrete adjacency matrix. Nonetheless, this approach can be computationally intensive with a growing number of nodes. In contrast, MTGNN \citep{wu2020connecting} and GDN \citep{deng2021graph} utilize a randomly initialized node embedding matrix to infer the latent graph structure. While this approach is less taxing on computational resources, it might compromise the accuracy of predictions.

\section{Methods}
\begin{figure}[t]
\centerline{\includegraphics[width=\textwidth]{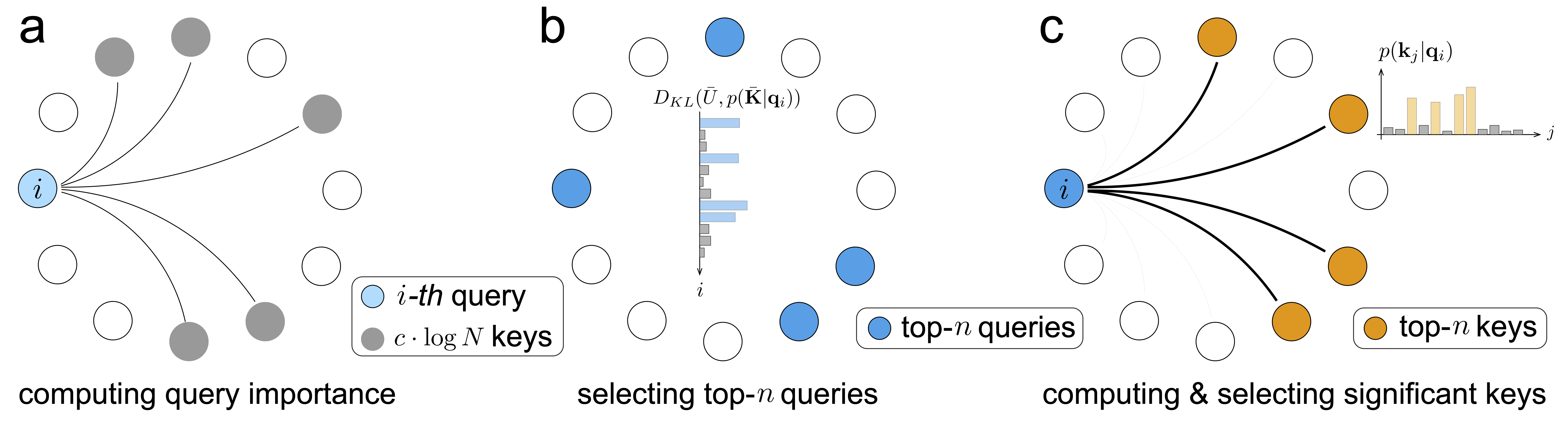}}
	\caption{\textbf{Overview of the latent graph structure learning (L-GSL).} \textbf{(a)} Key nodes chosen at random (depicted as gray circles) are used to measure the significance of a query node (shown as a blue circle). \textbf{(b)} Top-$n$ query nodes (blue circles) are picked according to the importance distribution across all query nodes. \textbf{(c)} Key nodes, colored in orange, that hold sufficient relevance to be linked with the chosen query node.}
	\label{fig1}
\end{figure}

In this section, we detail our proposed method, HGMTS. The overarching framework and core operational principles of this approach can be viewed in Figures \ref{fig1} and \ref{fig2}.

\subsection{Preliminaries}
Let $\bX \in \mathbb{R}^{N \times T \times M}$ represents a multivariate time series, where $N$ signifies the count of signals originating from various sensors, $T$ denotes the length of the sequence, and $M$ represents the dimension of the signal input (usually $M$ = 1). We depict this multivariate time series as a graph $\mathcal{G} = \{\mathcal{V},\mathcal{E},\mathcal{A}\}$, wherein the collection of nodes denoted by $\mathcal{V}$ corresponds to the sensors, the set $\mathcal{E}$ pertains to the edges, and $\mathcal{A}$ represents the adjacency matrix. Notably, the precise composition of $\mathcal{E}$ and $\mathcal{A}$ is not known initially; however, our model will acquire this knowledge through the learning process.

\subsection{Latent Graph Structure Learning (L-GSL)}\label{GSL}
We embrace the concept of self-attention (introduced by \citep{vaswani2017attention}) and employ the attention scores in the role of edge weights. The process of learning the adjacency matrix of the graph, denoted as $\mathcal{A}\in\mathbb{R}^{N\times N}$ unfolds as follows:
\begin{align}\label{eq1}
\bQ=\bH \mathbf{W}^Q,\; \bK=\bH\bW^K,\; \mathcal{A}=\operatorname{softmax}\left(\frac{\bQ \bK^T}{\sqrt{D}}\right)
\end{align}
where $\bH\in\mathbb{R}^{N\times D}$ corresponds to node embeddings\footnote{The methodology for computing node embeddings from multivariate time series is detailed in Section \ref{sigdecomp}}, $\mathbf{W}^Q \in\mathbb{R}^{D\times D}$ and $\mathbf{W}^K \in\mathbb{R}^{D\times D}$ are weight matrices that project $\bH$ into query $\mathbf{Q}$ and key $\mathbf{K}$, respectively. The main limitation in estimating latent graph structures in Eq.(\ref{eq1}) for a large value of $N$ is the necessity to perform quadratic time dot-product computations along with the utilization of $\mathcal{O}(N^2)$ memory. In an effort to achieve a self-attention mechanism complexity of $\mathcal{O}(N\log N)$, our approach involves identifying pivotal query nodes and their associated significant key nodes in a sequential manner.

\subsubsection{Identifying pivotal query nodes} For the purpose of determining which query nodes will establish connections with other nodes, our initial step involves evaluating the significance of queries. Recent studies \citep{child2019generating,li2019enhancing,beltagy2020longformer,zhou2021informer} have highlighted the existence of sparsity in the distribution of self-attention probabilities. Drawing inspiration from these findings, we establish the importance of queries based on the Kullback-Leibler (KL) divergence between a uniform distribution and the attention probability distribution of query nodes.

Let $\bq_i$ and $\bk_i$ represent the $i$-th row in matrices $\bQ$ and $\bK$ respectively. For a given query node, $p(\bk_j|\bq_i)=\exp(\bq_i\bk_j^{\top})/\sum_{\ell}\exp(\bq_i\bk_{\ell}^{\top})$ denotes the attention probability of the $i$-th query towards the $j$-th key node. Then, $p(\bK | \bq_i)=[p(\bk_1 | \bq_i) \ldots p(\bk_N | \bq_i)]$ indicates the probability distribution of how the $i$-th query allocates its attention/weight across all nodes. In this context, $D_{KL}(U, p(\bK | \bq_i))$ quantifies the deviation of a query node's attention probabilities from a uniform distribution $\mathcal{U}\{1,N\}$. This divergence measurement serves as a metric for identifying significant query nodes; a higher KL divergence suggests that a query's attention is mainly directed towards particular key nodes, rather than being evenly distributed. As a result, these query nodes are postulated to be suitable candidates for establishing sparse connections.

The traversal of all query nodes for this measurement, however, still entails a quadratic computational requirement. It is worth noting that a recent study demonstrated that the relative magnitudes of query importance remain unchanged even when the divergence metric is calculated using randomly sampled keys \citep{zhou2021informer}. Building on this idea, we determine the importance of query nodes through the computation of $D_{KL}(\bar{U}, p(\bar{\mathbf{K}} | \mathbf{q}_i))$ instead, where $\bar U=\mathcal{U}\{1,n\}$, $\bar\bK$ represents a matrix containing randomly sampled $n$ row vectors from $\mathbf{K}$, and $n=\lfloor c\cdot\log N\rfloor$ denotes the number of random samples based on a constant sampling factor $c$ (Figure \ref{fig1}a). Given this measurement of query importance, we select top-$n$ query nodes and denote it as $\bar\bQ$ (Figure \ref{fig1}b).

\subsubsection{Identifying associated key nodes}
Using the selected set of $n$ query nodes, our subsequent step involves identifying the corresponding key nodes to establish connections. In pursuit of achieving this objective, we initiate by computing the attention probabilities $p(\bK|\bq_i)$ of the $i$-th query across all keys nodes; this procedure is reiterated for each of the $n$ query nodes. Next, we choose the top-$n$ key nodes for each query based on their attention scores (Figure \ref{fig1}c), and we designate this collection as $\bar\bK$. The ultimate adjacency matrix, adhering to the sparsity constraint, is defined by the equation:
\begin{align}\label{eq2}
\bar{\mathcal{A}}=\operatorname{softmax}\left(\frac{\bar\bQ \bar\bK^T}{\sqrt{D}}\right)
\end{align}
In this equation, $\bar\bQ$ and $\bar\bK$ possess the same dimensions as $\bQ$ and $\bK$, except that the row vectors corresponding to insignificant query and key nodes are replaced with zeros. To sum up, the complexity of all the necessary computations for evaluating the significance of a query node and determining which key nodes to establish connections with, considering the top-$n$ chosen queries, amounts to $\mathcal{O}(N \log N)$.

\subsection{Hierarchical Signal Decomposition}\label{sigdecomp}
\begin{figure}[t]
\centerline{\includegraphics[width=\textwidth]{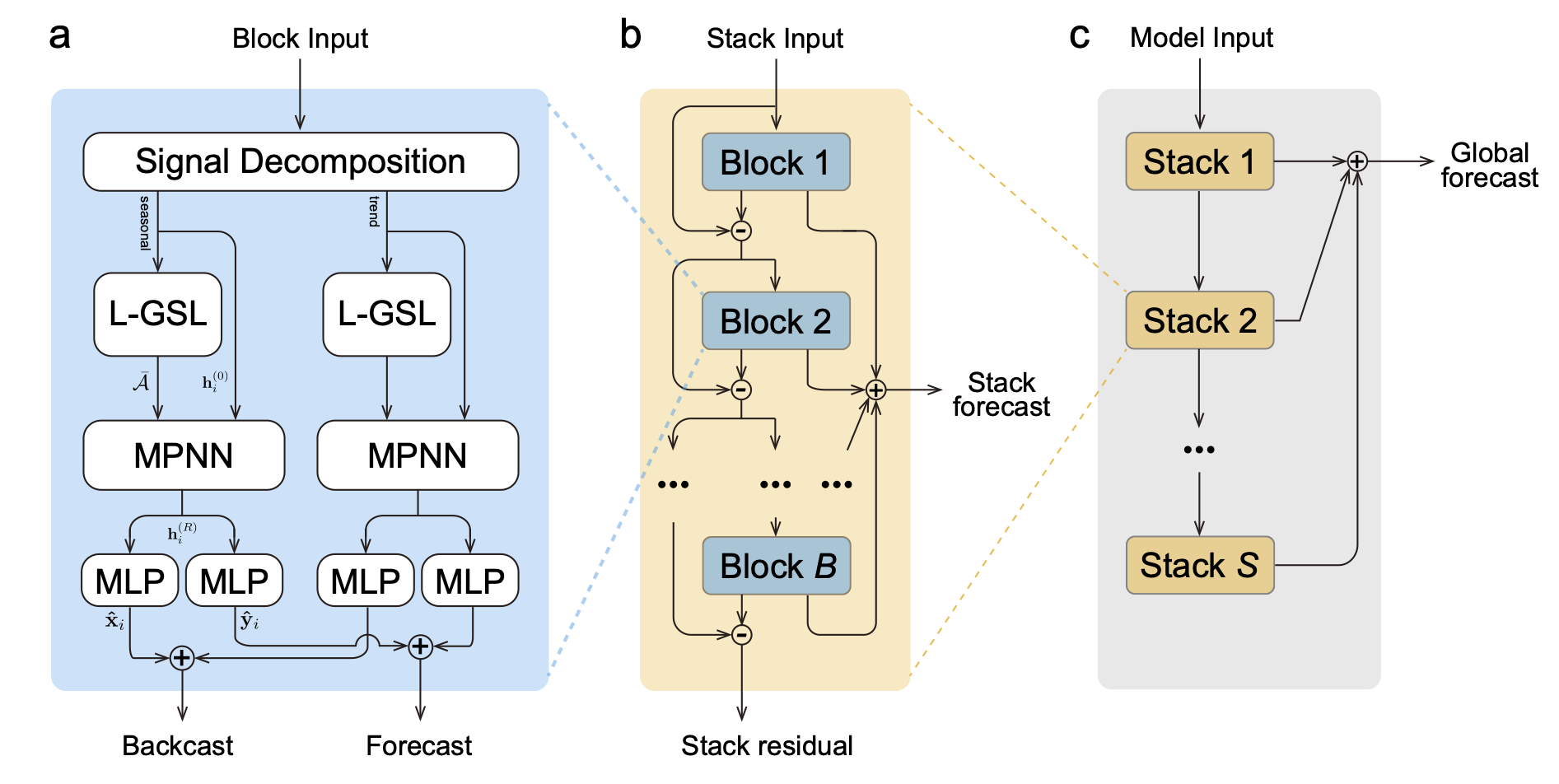}}
	\caption{\textbf{Overview of the proposed HGMTS model architecture.} \textbf{(a)} The hierarchical residual block is marked by signal decomposition and GNN-centric L-GSL modules. \textbf{(b)} The combination of multiple blocks forms a stack, \textbf{(c)} culminating in the entire model design to ultimately produce a global forecasting output.}
	\label{fig2}
\end{figure}

This section provides an overview of the proposed approach shown in Figure \ref{fig2} and discusses the overall design principles. Our approach builds upon N-BEATS \citep{oreshkin2020N-BEATS}, enhancing its key elements significantly. Our main methodology comprises of three primary elements: signal decomposition, latent graph structure learning, and constructing forecasts and backcasts in a hierarchical manner.
Much like the N-BEATS approach, every block is trained to generate signals for both backcast and forecast outputs. Here, the backcast output is designed to be subtracted from the input of the subsequent block, whereas the forecasts are combined to produce the final prediction (Figure \ref{fig2}). These blocks are arranged in stacks, each focusing on a distinct spatial dependency through a unique set of graph structures.

\subsubsection{Signal decomposition module}
Recent research has witnessed a surging interest in disentangling time series data into its trend and seasonal components. These components respectively represent the overall long-term pattern and the seasonal fluctuations within the time signals. However, when it comes to future time series, directly performing this decomposition becomes impractical due to the inherent uncertainty of the future. To address this challenge, we propose the incorporation of a signal decomposition module within a single block (Figure \ref{fig2}a). This module enables the gradual extraction of the consistent, long-term trend from intermediate forecasting signals. Specifically, we employ the moving average technique to smooth out recurring fluctuations and uncover the underlying long-term trends as outlined below:
\begin{equation}\label{eq2}
\bX^{\texttt{trend}}=\operatorname{AvgPool}(\operatorname{Padding}(\bX)),\;\;\; \bX^{{\texttt{seas}}} = \bX - \bX^{\texttt{trend}}
\end{equation}
where $\bX^{{\texttt{trend}}}, \bX^{{\texttt{seas}}}$ denote the trend and seasonal components respectively. We opt for the $\operatorname{AvgPool}(\cdot)$ for the moving average, accompanied by the zero padding operation to maintain the original series length intact.

\subsubsection{Message-passing module}
The message-passing module receives as input the past $L$ time steps of both seasonal and trend outputs $\bX^{{\texttt{seas}}}_{t-L:t}, \bX^{{\texttt{trend}}}_{t-L:t}\in\mathbb{R}^{N\times L}$ obtained from the signal decomposition. As the two components go through the same set of distinct parameterized network modules, their differentiation will be disregarded henceforth. At each time step $t$, the input consisting of $N$ multivariate time series with $L$ lags are transformed into embedding vectors $\bH\in\mathbb{R}^{N\times D}$ using a multilayer perceptron (MLP). Each row vector $\bh_i$ in this matrix represents an individual node embedding. Subsequently, these node embeddings are employed in Eq.\ref{eq1} of the latent graph structure learning module to create a sparse adjacency matrix $\mathcal{\bar A}$. This matrix, in conjunction with the node embedding matrix, serves as the input for the message-passing neural network (Figure \ref{fig2}a). To be more specific, the $r$-th round of message passing in the GNN is executed using the following equations:\newline
\begin{minipage}{.4\linewidth}
    \centering
    \begin{align}
        \bh_i^{(0)} &= f(\bx_{i,t-L:t})\\
        \bm_{ij}^{(r)} &= g(\bh_i^{(r)}-\bh_j^{(r)})
    \end{align}
\end{minipage}
\begin{minipage}{.58\linewidth}
    \centering
    \begin{align}
        \mathcal{\bar A} &= \text{L-GSL}(\bH)\\
        \bh_i^{(r+1)} &= \text{GRU}(\bh_i^{(r)}, \textstyle\sum\nolimits_{j \in \mathcal{N}(i)} \bar a_{ij} \cdot\bm_{i j}^{(r)})
    \end{align}
\end{minipage}

where $\bh_i^{(r)}$ refers to the $i$-th node embedding after round $r$, and $\bm_{ij}^{(r)}$ represents the message vector from node $i$ to $j$. The interaction strength associated with the edge $(i,j)$, denoted as $\bar a_{ij}$, corresponds to the entry in $\mathcal{\bar A}$ at the $i$-th row and $j$-th column. Both the encoding function $f(\cdot)$ and the message function $g(\cdot)$ are implemented as two-layer MLPs with ReLU nonlinearities. Finally, the node embeddings are updated using a GRU after aggregating all incoming messages through a weighted sum over the neighborhood $\mathcal{N}(i)$ for each node $i$. This sequence of operations is repeated separately for the seasonal and trend inputs, with no sharing of parameters (Figure \ref{fig2}a).

To enhance both the model's expressivity and its capacity for generalization, we employ a multi-module GNN framework \citep{lee2023towards}. More specifically, the next hidden state $\bh_{i}^{(r+1)}$ is computed by blending two intermediate node states, $\bh_{i,1}^{(r)}$ and $\bh_{i,2}^{(r)}$, through a linear combination defined as follows:
\begin{eqnarray}\label{eq:gnn_update}
  \bh_{i}^{(r+1)} &=\; \beta_i^{(r)}\bh_{i,1}^{(r)} + (1-\beta_i^{(r)})\bh_{i,2}^{(r)}
\end{eqnarray}
where the two intermediate representations $\bh_{i,1}^{(r)}$ and $\bh_{i,2}^{(r)}$ are derived from Eq. (7) using two distinct GRUs. The value of the gating variable $\beta_i^{(r)}$ is determined by another processing unit employing a gating function $\xi_g$, which is a neural network producing a scalar output through a sigmoid activation.

\subsubsection{Forecast and backcast module}
Following the completion of the last $R$ round of message passing (3 rounds in total), the backcast $\hat\bx$ and forecast outputs $\hat\by$ are generated in this procedure. This is achieved by mapping the final node embeddings through separate two MLPs. These MLPs are responsible for handling the generation of backcast and forecast outputs individually (Figure \ref{fig2}a). It is important to note that the last layer of these MLPs is designed as a linear layer. This process of generating backcast and forecast outputs is applied to both the seasonal and trend pathways, and the ultimate backcast and forecast outputs are obtained by summing up the respective outputs from the seasonal and trend components (Figure \ref{fig2}a):\newline
\begin{minipage}{.48\linewidth}
    \centering
    \begin{align}
        \hat\bx_{i,t-L:t}^{\texttt{seas}} &= \phi_{\texttt{seas}}(\bh_{i,\texttt{seas}}^{(R)}) \nonumber\\
        \hat\bx_{i,t-L:t}^{\texttt{trend}} &= \phi_{\texttt{trend}}(\bh_{i,\texttt{trend}}^{(R)}) \nonumber\\
        \hat\bx_{i,t-L:t} &= \hat\bx_{i,t-L:t}^{\texttt{seas}} + 
    \hat\bx_{i,t-L:t}^{\texttt{trend}} \nonumber
    \end{align}
\end{minipage}
\begin{minipage}{.48\linewidth}
    \centering
    \begin{align}
        \hat\by_{i,t+1:t+K}^{\texttt{seas}} &= \psi_{\texttt{seas}}(\bh_{i,\texttt{seas}}^{(R)}) \\
        \hat\by_{i,t+1:t+K}^{\texttt{trend}} &= \psi_{\texttt{trend}}(\bh_{i,\texttt{trend}}^{(R)}) \\
        \hat\by_{i,t+1:t+K} &= \hat\by_{i,t+1:t+K}^{\texttt{seas}} + 
    \hat\by_{i,t+1:t+K}^{\texttt{trend}} 
    \end{align}
\end{minipage}

Here, $\phi_{{}_\square}$ and $\psi_{{}_\square}$ represent two-layer MLPs designed to acquire the predictive decomposition of the partial backcast $\hat\bx_{i,t-L:t}$ of the preceding $L$ time steps, and the forecast $\hat\by_{i,t+1:t+K}$ of the subsequent $K$ time steps. These MLPs operate on components denoted as $\square$, which can be either the \texttt{seasonal} or \texttt{trend} aspects. Note that the indexing related to block or stack levels has been excluded for clarity. The resulting global forecast is constructed by summing the outputs of all blocks (Figure \ref{fig2}b-c).

\section{Experimental Setup}
We first provide an overview of the datasets (Table \ref{tab1}), evaluation metrics, and baselines employed to quantitatively assess our model's performance. The main results are summarized in Table \ref{tab2}, demonstrating the competitive predictive performance of our approach in comparison to existing works. We then elaborate on the specifics of our training and evaluation setups followed by detailing the ablation studies.

\subsection{Datasets} 
Our experimentation extensively covers six real-world benchmark datasets. Conforming to the standard protocol \citep{wu2021autoformer,zhou2021informer}, the split of all datasets into training, validation, and test sets has been conducted chronologically, following a split ratio of 60:20:20 for the ETTm$_2$ dataset and a split ratio of 70:10:20 for the remaining datasets.
\begin{itemize}
\item $\textbf{ETTm}_2$ \textbf{(Electricity Transformer Temperature)}: This dataset encompasses data obtained from electricity transformers, featuring load and oil temperatures recorded every 15 minutes during the period spanning from July 2016 to July 2018.
\item \textbf{ECL (Electricity Consuming Load)}: The ECL dataset compiles hourly electricity consumption (in Kwh) data from 321 customers, spanning the years 2012 to 2014.
\item \textbf{Exchange}: This dataset aggregates daily exchange rates of eight different countries relative to the US dollar. The data spans from 1990 to 2016.
\item \textbf{Traffic}: The Traffic dataset is a collection of road occupancy rates from 862 sensors situated along San Francisco Bay area freeways. These rates are recorded every hour, spanning from January 2015 to December 2016.
\item \textbf{Weather}: This dataset comprises 21 meteorological measurements, including air temperature and humidity. These measurements are recorded every 10 minutes throughout the entirety of the year 2020 in Germany.
\item \textbf{ILI (Influenza-Like Illness)}: This dataset provides a record of weekly influenza-like illness (ILI) patients and the total patient count, sourced from the Centers for Disease Control and Prevention of the US. The data covers the extensive period from 2002 to 2021. It represents the ratio of ILI patients versus the total count for each week.
\end{itemize}

\subsection{Evaluation metrics}
We evaluate the effectiveness of our approach by measuring its accuracy using the mean squared error (MSE) and mean absolute error (MAE) metrics. These evaluations are conducted for various prediction horizon lengths $K\in\{96, 192, 336, 720\}$ given a fixed input length $L=96$, except for ILI where $L=36$:
\begin{equation}
\mathrm{MSE}=\frac{1}{NK} \sum_{i=1}^N\sum_{\tau=t}^{t+K}\left(\mathbf{y}_{i,\tau}-\hat{\mathbf{y}}_{i,\tau}\right)^2, \quad \mathrm{MAE}=\frac{1}{NK} \sum_{i=1}^N\sum_{\tau=t}^{t+K}\left|\mathbf{y}_{i,\tau}-\hat{\mathbf{y}}_{i,\tau}\right|
\end{equation}

\subsection{Baselines}
We evaluate our proposed model by comparing it with seven baseline models. These include: (1) N-BEATS \citep{oreshkin2020N-BEATS}, which aligns with the external structure of our model, (2) Autoformer \citep{wu2021autoformer}, (3) Informer \citep{zhou2021informer}, (4) Reformer \citep{Kitaev2020Reformer}, (5) LogTrans \citep{li2019enhancing} -- latest transformer-based models. Additionally, we compare with two conventional RNN-based models: (6) LSTNet \citep{lai2018modeling} and (7) LSTM \citep{hochreiter1997long}.

\subsection{Hyperparameters}
Our model is trained using the ADAM optimizer, starting with a learning rate of $10^{-4}$ that gets reduced by half every two epochs. We employ early stopping during training, stopping the process if there is no improvement after 10 epochs. The training is carried out with a batch size of 32. We have configured our model with 3 stacks, each containing 1 block. All tests are conducted three times, making use of the PyTorch framework, and are executed on a single NVIDIA RTX 3090 with 24GB GPU.

\section{Experimental Results}

\subsection{Multivariate time series forecasting}
In the multivariate setting, our proposed model, HGMTS, consistently achieves state-of-the-art performance across all benchmark datasets and prediction length configurations (Table \ref{tab2}). Notably, under the input-96-predict-192 setting, HGMTS demonstrates significant improvements over previous state-of-the-art results, with a 34\% (0.273→0.180) reduction in MSE for ETT, 19\% (0.180→0.146) reduction for ECL, 53\% (0.225→0.105) reduction for Exchange, 5\% (0.409→0.389) reduction for Traffic, and 10\% (0.229→0.207) reduction for Weather. In the case of the input-36-predict-60 setting for ILI, HGMTS achieves 17\% (2.547→2.118) reduction in MSE. Overall, HGMTS delivers an average MSE reduction of 23\% across these settings. It is particularly striking how HGMTS drastically improves predictions for the Exchange dataset, where it records an average MSE reduction of 52\% for all prediction lengths. Moreover, HGMTS stands out for its outstanding long-term stability, an essential attribute for real-world applications.

\subsection{Effect of sparsity in graphs on forecasting}
 Within the HGMTS model framework, a key hyperparameter is the sampling factor in L-GSL. This factor determines how many query nodes are selected and subsequently linked to key nodes. For the sake of simplicity, we ensure that the number of chosen query and key nodes remains the same. We then measure the sparsity of the latent graphs by computing the proportion of selected pivotal query or key nodes relative to the total time series count. This proportion is denoted as $\gamma=\lfloor c\cdot\log N\rfloor/N$ and acts as an indicator of the sparsity in building these latent graphs.

To understand the impact of sparsity in the learned graphs, we modify $\gamma$ values between 0.2 and 0.7 and then document the findings from the multivariate forecasting studies. As detailed in Table \ref{tab2}, there is a consistent trend: all the graphs lean towards sparse interactions ($\gamma \le 0.5$), targeting optimal predictive outcomes in LSTF tasks. Additionally, different benchmark datasets exhibit unique preferences regarding the optimal sparsity level for predictive performance, as displayed in Table \ref{tab2}.

\begin{figure}[t]
\vspace{-0.2in}
\centerline{\includegraphics[width=\textwidth]{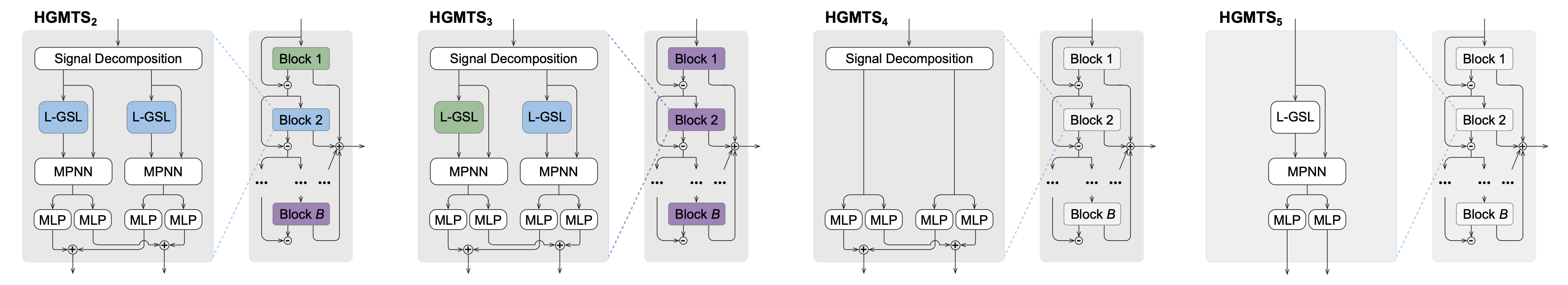}}
	\caption{\textbf{Ablation study overview.} Displayed are four distinct model architectures explored to understand the impact of specific components on overall LSTF performance.}
	\label{fig3}
\end{figure}

\subsection{Ablation studies}
We posit that the strengths of the HGMTS architecture stem from its ability to hierarchically model the interplay between time series, particularly in the realms of trend and seasonality components.  To delve deeper into this proposition, we present a series of control models for a comparative analysis:

\begin{itemize}
    \item HGMTS$_1$: The model as showcased in Figure \ref{fig2}.
    \item HGMTS$_2$: A model that has shared latent graphs between trend and seasonality channels, but not across different blocks and stacks.
    \item HGMTS$_3$: A model where latent graphs are shared throughout all blocks and stacks but remain distinct between trend and seasonality channels.
    \item HGMTS$_4$: This model omits the L-GSL and MPNN modules.
    \item HGMTS$_5$: A model focusing solely on either the trend or seasonality channel, essentially lacking the signal decomposition module.
    \item HGMTS$_6$: A model that has used a single GRU module in Eq (\ref{eq:gnn_update}).
\end{itemize}

Under the same multivariate setting, the evaluation metrics for each control model, averaged over all benchmark datasets excluding ILI, are detailed in Table \ref{tab4}. The HGMTS$_4$, which forgoes the L-GSL and MPNN modules, experiences a noticeable average MSE surge of 30\% (0.258→0.336) across all horizons. This rise is the most significant among all controls, indicating that capturing interdependencies between multivariate signals is vital in our suggested model. HGMTS$_5$, which emphasizes solely on a single channel between trend and seasonality, registers the second most pronounced MSE growth (18\%: 0.258→0.305), suggesting that signal decomposition is also instrumental in LSTF tasks. Sharing the latent graphs -- whether between the trend and seasonality pathways (as in HGMTS$_2$) or among blocks (as in HGMTS$_3$) -- does elevate the average MSE, but the rise is modest when compared with the first two control models. Additionally, our findings highlight that incorporating multiple node update mechanisms in MPNN, as seen in HGMTS$_6$, brings about a slight enhancement in forecasting precision.

The information presented in Table \ref{tab4} robustly supports the idea that best performance is achieved by integrating both suggested components: the latent graph structure and hierarchical signal decomposition. This emphasizes their synergistic role in enhancing the accuracy of long sequence time series predictions. Furthermore, it is confirmed that crafting distinct latent associations between time series hierarchically, spanning both trend and seasonal channels, is instrumental in attaining improved prediction outcomes.

\section{Conclusions}
In this paper, we delved into the challenge of long-term multivariate time series forecasting, an area that has seen notable progress recently. However, the intricate temporal patterns often impede models from effectively learning reliable dependencies. In response, we introduce HGMTS, a spatio-temporal multivariate time series forecasting model that incorporates a signal decomposition module and employs a latent graph structure learning as intrinsic operators. This unique approach allows for the hierarchical aggregation of long-term trend and seasonal information from intermediate predictions. Furthermore, we adopt a multi-module message-passing framework to enhance our model's capacity to capture diverse time series data from a range of heterogeneous sensors. This approach distinctly sets us apart from previous neural forecasting models. Notably, HGMTS naturally achieves a computational complexity of $\mathcal{O}(N\log N)$ and consistently delivers state-of-the-art performance across a wide array of real-world datasets.

Learning a latent graph typically poses considerable challenges. Even though our model leverages the top-k pooling method to infer the latent graph, there are many other deep learning techniques that could be investigated in upcoming studies to uncover hidden structural patterns. Enhancements related to both representation capacity and computational efficiency might expand its broader adoption.

\section*{Acknowledgments}
This work was supported in part by the National Research Foundation of Korea (NRF) grant (No. NRF-2021R1F1A1045390), the Brain Convergence Research Program (No. NRF-2021M3E5D2A01023887), the Bio \& Medical Technology Development Program (No. RS-2023-00226494) of the National Research Foundation (NRF), the Institute of Information \& communications Technology Planning \& Evaluation (IITP) grant (No.2020-0-01373, Artificial Intelligence Graduate School Program (Hanyang University)) funded by the Korean government (MSIT), the Technology Innovation Program (20013726, Development of Industrial Intelligent Technology for Manufacturing, Process, and Logistics) funded By the Ministry of Trade, Industry \& Energy (MOTIE, Korea), and in part by Samsung Electronics Co., Ltd.

\newpage
\bibliographystyle{unsrtnat}
\bibliography{references}

\begin{thebibliography}{35}
\providecommand{\natexlab}[1]{#1}
\providecommand{\url}[1]{\texttt{#1}}
\expandafter\ifx\csname urlstyle\endcsname\relax
  \providecommand{\doi}[1]{doi: #1}\else
  \providecommand{\doi}{doi: \begingroup \urlstyle{rm}\Url}\fi

\bibitem[Petropoulos et~al.(2022)Petropoulos, Apiletti, Assimakopoulos, Babai,
  Barrow, Taieb, Bergmeir, Bessa, Bijak, Boylan,
  et~al.]{petropoulos2022forecasting}
Fotios Petropoulos, Daniele Apiletti, Vassilios Assimakopoulos, Mohamed~Zied
  Babai, Devon~K Barrow, Souhaib~Ben Taieb, Christoph Bergmeir, Ricardo~J
  Bessa, Jakub Bijak, John~E Boylan, et~al.
\newblock Forecasting: theory and practice.
\newblock \emph{International Journal of Forecasting}, 2022.

\bibitem[Lam et~al.(2022)Lam, Sanchez-Gonzalez, Willson, Wirnsberger,
  Fortunato, Pritzel, Ravuri, Ewalds, Alet, Eaton-Rosen,
  et~al.]{lam2022graphcast}
Remi Lam, Alvaro Sanchez-Gonzalez, Matthew Willson, Peter Wirnsberger, Meire
  Fortunato, Alexander Pritzel, Suman Ravuri, Timo Ewalds, Ferran Alet, Zach
  Eaton-Rosen, et~al.
\newblock Graphcast: Learning skillful medium-range global weather forecasting.
\newblock \emph{arXiv preprint arXiv:2212.12794}, 2022.

\bibitem[Derrow-Pinion et~al.(2021)Derrow-Pinion, She, Wong, Lange, Hester,
  Perez, Nunkesser, Lee, Guo, Wiltshire, et~al.]{derrow2021eta}
Austin Derrow-Pinion, Jennifer She, David Wong, Oliver Lange, Todd Hester, Luis
  Perez, Marc Nunkesser, Seongjae Lee, Xueying Guo, Brett Wiltshire, et~al.
\newblock Eta prediction with graph neural networks in google maps.
\newblock In \emph{Proceedings of the 30th ACM International Conference on
  Information \& Knowledge Management}, pages 3767--3776, 2021.

\bibitem[Rahmani et~al.(2023)Rahmani, Baghbani, Bouguila, and
  Patterson]{rahmani2023graph}
Saeed Rahmani, Asiye Baghbani, Nizar Bouguila, and Zachary Patterson.
\newblock Graph neural networks for intelligent transportation systems: A
  survey.
\newblock \emph{IEEE Transactions on Intelligent Transportation Systems}, 2023.

\bibitem[Wu et~al.(2020)Wu, Pan, Long, Jiang, Chang, and
  Zhang]{wu2020connecting}
Zonghan Wu, Shirui Pan, Guodong Long, Jing Jiang, Xiaojun Chang, and Chengqi
  Zhang.
\newblock Connecting the dots: Multivariate time series forecasting with graph
  neural networks.
\newblock In \emph{Proceedings of the 26th ACM SIGKDD international conference
  on knowledge discovery \& data mining}, pages 753--763, 2020.

\bibitem[Shang et~al.(2021)Shang, Chen, and Bi]{shang2021discrete}
Chao Shang, Jie Chen, and Jinbo Bi.
\newblock Discrete graph structure learning for forecasting multiple time
  series.
\newblock \emph{arXiv preprint arXiv:2101.06861}, 2021.

\bibitem[Battaglia et~al.(2016)Battaglia, Pascanu, Lai, Rezende,
  et~al.]{battaglia2016interaction}
Peter Battaglia, Razvan Pascanu, Matthew Lai, Danilo~Jimenez Rezende, et~al.
\newblock Interaction networks for learning about objects, relations and
  physics.
\newblock In \emph{Advances in Neural Information Processing Systems}, pages
  4502--4510, 2016.

\bibitem[Kipf et~al.(2018)Kipf, Fetaya, Wang, Welling, and
  Zemel]{kipf2018neural}
Thomas Kipf, Ethan Fetaya, Kuan-Chieh Wang, Max Welling, and Richard Zemel.
\newblock Neural relational inference for interacting systems.
\newblock In \emph{International Conference on Machine Learningachine
  learning}, pages 2688--2697. PMLR, 2018.

\bibitem[Sanchez-Gonzalez et~al.(2020)Sanchez-Gonzalez, Godwin, Pfaff, Ying,
  Leskovec, and Battaglia]{sanchez2020learning}
Alvaro Sanchez-Gonzalez, Jonathan Godwin, Tobias Pfaff, Rex Ying, Jure
  Leskovec, and Peter Battaglia.
\newblock Learning to simulate complex physics with graph networks.
\newblock In \emph{International Conference on Machine Learning}, pages
  8459--8468. PMLR, 2020.

\bibitem[Deng and Hooi(2021)]{deng2021graph}
Ailin Deng and Bryan Hooi.
\newblock Graph neural network-based anomaly detection in multivariate time
  series.
\newblock In \emph{Proceedings of the AAAI Conference on Artificial
  Intelligence}, volume~35, pages 4027--4035, 2021.

\bibitem[Cleveland et~al.(1990)Cleveland, Cleveland, McRae, and
  Terpenning]{cleveland1990stl}
Robert~B Cleveland, William~S Cleveland, Jean~E McRae, and Irma Terpenning.
\newblock Stl: A seasonal-trend decomposition.
\newblock \emph{Journal of Official Statistics}, 6\penalty0 (1):\penalty0
  3--73, 1990.

\bibitem[Oreshkin et~al.(2020)Oreshkin, Carpov, Chapados, and
  Bengio]{oreshkin2020N-BEATS}
Boris~N. Oreshkin, Dmitri Carpov, Nicolas Chapados, and Yoshua Bengio.
\newblock N-beats: Neural basis expansion analysis for interpretable time
  series forecasting.
\newblock In \emph{International Conference on Learning Representations}, 2020.

\bibitem[Challu et~al.(2023)Challu, Olivares, Oreshkin, Ramirez, Canseco, and
  Dubrawski]{challu2023nhits}
Cristian Challu, Kin~G Olivares, Boris~N Oreshkin, Federico~Garza Ramirez,
  Max~Mergenthaler Canseco, and Artur Dubrawski.
\newblock Nhits: Neural hierarchical interpolation for time series forecasting.
\newblock In \emph{Proceedings of the AAAI Conference on Artificial
  Intelligence}, volume~37, pages 6989--6997, 2023.

\bibitem[Yu et~al.(2017)Yu, Zheng, Anandkumar, and Yue]{yu2017long}
Rose Yu, Stephan Zheng, Anima Anandkumar, and Yisong Yue.
\newblock Long-term forecasting using tensor-train rnns.
\newblock \emph{Arxiv}, 2017.

\bibitem[Qin et~al.(2017)Qin, Song, Chen, Cheng, Jiang, and
  Cottrell]{qin2017dual}
Yao Qin, Dongjin Song, Haifeng Chen, Wei Cheng, Guofei Jiang, and Garrison
  Cottrell.
\newblock A dual-stage attention-based recurrent neural network for time series
  prediction.
\newblock \emph{International Joint Conference on Artificial Intelligence},
  2017.

\bibitem[Wen et~al.(2017)Wen, Torkkola, Narayanaswamy, and
  Madeka]{wen2017multi}
Ruofeng Wen, Kari Torkkola, Balakrishnan Narayanaswamy, and Dhruv Madeka.
\newblock A multi-horizon quantile recurrent forecaster.
\newblock \emph{arXiv preprint arXiv:1711.11053}, 2017.

\bibitem[Salinas et~al.(2020)Salinas, Flunkert, Gasthaus, and
  Januschowski]{salinas2020deepar}
David Salinas, Valentin Flunkert, Jan Gasthaus, and Tim Januschowski.
\newblock Deepar: Probabilistic forecasting with autoregressive recurrent
  networks.
\newblock \emph{International Journal of Forecasting}, 36\penalty0
  (3):\penalty0 1181--1191, 2020.

\bibitem[Sutskever et~al.(2014)Sutskever, Vinyals, and
  Le]{sutskever2014sequence}
Ilya Sutskever, Oriol Vinyals, and Quoc~V Le.
\newblock Sequence to sequence learning with neural networks.
\newblock \emph{Advances in neural information processing systems}, 27, 2014.

\bibitem[Beltagy et~al.(2020)Beltagy, Peters, and Cohan]{beltagy2020longformer}
Iz~Beltagy, Matthew~E Peters, and Arman Cohan.
\newblock Longformer: The long-document transformer.
\newblock \emph{arXiv preprint arXiv:2004.05150}, 2020.

\bibitem[Kitaev et~al.(2020)Kitaev, Kaiser, and Levskaya]{Kitaev2020Reformer}
Nikita Kitaev, Lukasz Kaiser, and Anselm Levskaya.
\newblock Reformer: The efficient transformer.
\newblock In \emph{International Conference on Learning Representations}, 2020.
\newblock URL \url{https://openreview.net/forum?id=rkgNKkHtvB}.

\bibitem[Zhou et~al.(2021)Zhou, Zhang, Peng, Zhang, Li, Xiong, and
  Zhang]{zhou2021informer}
Haoyi Zhou, Shanghang Zhang, Jieqi Peng, Shuai Zhang, Jianxin Li, Hui Xiong,
  and Wancai Zhang.
\newblock Informer: Beyond efficient transformer for long sequence time-series
  forecasting.
\newblock In \emph{Proceedings of the AAAI Conference on Artificial
  Intelligence}, volume~35, pages 11106--11115, 2021.

\bibitem[Chen et~al.(2021)Chen, Peng, Fu, and Ling]{chen2021autoformer}
Minghao Chen, Houwen Peng, Jianlong Fu, and Haibin Ling.
\newblock Autoformer: Searching transformers for visual recognition.
\newblock In \emph{Proceedings of the IEEE/CVF international conference on
  computer vision}, pages 12270--12280, 2021.

\bibitem[Woo et~al.(2022)Woo, Liu, Sahoo, Kumar, and Hoi]{woo2022etsformer}
Gerald Woo, Chenghao Liu, Doyen Sahoo, Akshat Kumar, and Steven Hoi.
\newblock Etsformer: Exponential smoothing transformers for time-series
  forecasting.
\newblock \emph{arXiv preprint arXiv:2202.01381}, 2022.

\bibitem[Yan et~al.(2018)Yan, Xiong, and Lin]{yan2018spatial}
Sijie Yan, Yuanjun Xiong, and Dahua Lin.
\newblock Spatial temporal graph convolutional networks for skeleton-based
  action recognition.
\newblock In \emph{Proceedings of the AAAI conference on artificial
  intelligence}, volume~32, 2018.

\bibitem[Huang et~al.(2020)Huang, Shen, Tian, Li, Huang, and
  Hua]{huang2020spatio}
Zhen Huang, Xu~Shen, Xinmei Tian, Houqiang Li, Jianqiang Huang, and Xian-Sheng
  Hua.
\newblock Spatio-temporal inception graph convolutional networks for
  skeleton-based action recognition.
\newblock In \emph{Proceedings of the 28th ACM International Conference on
  Multimedia}, pages 2122--2130, 2020.

\bibitem[Li et~al.(2017)Li, Yu, Shahabi, and Liu]{li2017diffusion}
Yaguang Li, Rose Yu, Cyrus Shahabi, and Yan Liu.
\newblock Diffusion convolutional recurrent neural network: Data-driven traffic
  forecasting.
\newblock \emph{arXiv preprint arXiv:1707.01926}, 2017.

\bibitem[Seo et~al.(2018)Seo, Defferrard, Vandergheynst, and
  Bresson]{seo2018structured}
Youngjoo Seo, Micha{\"e}l Defferrard, Pierre Vandergheynst, and Xavier Bresson.
\newblock Structured sequence modeling with graph convolutional recurrent
  networks.
\newblock In \emph{Neural Information Processing: 25th International
  Conference, ICONIP 2018, Siem Reap, Cambodia, December 13-16, 2018,
  Proceedings, Part I 25}, pages 362--373. Springer, 2018.

\bibitem[Zhao et~al.(2019)Zhao, Song, Zhang, Liu, Wang, Lin, Deng, and
  Li]{zhao2019t}
Ling Zhao, Yujiao Song, Chao Zhang, Yu~Liu, Pu~Wang, Tao Lin, Min Deng, and
  Haifeng Li.
\newblock T-gcn: A temporal graph convolutional network for traffic prediction.
\newblock \emph{IEEE transactions on intelligent transportation systems},
  21\penalty0 (9):\penalty0 3848--3858, 2019.

\bibitem[Vaswani et~al.(2017)Vaswani, Shazeer, Parmar, Uszkoreit, Jones, Gomez,
  Kaiser, and Polosukhin]{vaswani2017attention}
Ashish Vaswani, Noam Shazeer, Niki Parmar, Jakob Uszkoreit, Llion Jones,
  Aidan~N Gomez, {\L}ukasz Kaiser, and Illia Polosukhin.
\newblock Attention is all you need.
\newblock In \emph{Advances in neural information processing systems}, pages
  5998--6008, 2017.

\bibitem[Child et~al.(2019)Child, Gray, Radford, and
  Sutskever]{child2019generating}
Rewon Child, Scott Gray, Alec Radford, and Ilya Sutskever.
\newblock Generating long sequences with sparse transformers.
\newblock \emph{arXiv preprint arXiv:1904.10509}, 2019.

\bibitem[Li et~al.(2019)Li, Jin, Xuan, Zhou, Chen, Wang, and
  Yan]{li2019enhancing}
Shiyang Li, Xiaoyong Jin, Yao Xuan, Xiyou Zhou, Wenhu Chen, Yu-Xiang Wang, and
  Xifeng Yan.
\newblock Enhancing the locality and breaking the memory bottleneck of
  transformer on time series forecasting.
\newblock \emph{Advances in Neural Information Processing Systems}, 32, 2019.

\bibitem[Lee and Yoon(2023)]{lee2023towards}
HyunGeun Lee and Kijung Yoon.
\newblock Towards better generalization with flexible representation of
  multi-module graph neural networks.
\newblock \emph{Transactions on Machine Learning Research}, 2023.
\newblock ISSN 2835-8856.
\newblock URL \url{https://openreview.net/forum?id=EYjfLeJL4l}.

\bibitem[Wu et~al.(2021)Wu, Xu, Wang, and Long]{wu2021autoformer}
Haixu Wu, Jiehui Xu, Jianmin Wang, and Mingsheng Long.
\newblock Autoformer: Decomposition transformers with auto-correlation for
  long-term series forecasting.
\newblock \emph{Advances in Neural Information Processing Systems},
  34:\penalty0 22419--22430, 2021.

\bibitem[Lai et~al.(2018)Lai, Chang, Yang, and Liu]{lai2018modeling}
Guokun Lai, Wei-Cheng Chang, Yiming Yang, and Hanxiao Liu.
\newblock Modeling long-and short-term temporal patterns with deep neural
  networks.
\newblock In \emph{The 41st international ACM SIGIR conference on research \&
  development in information retrieval}, pages 95--104, 2018.

\bibitem[Hochreiter and Schmidhuber(1997)]{hochreiter1997long}
Sepp Hochreiter and J{\"u}rgen Schmidhuber.
\newblock Long short-term memory.
\newblock \emph{Neural Computation}, 9\penalty0 (8):\penalty0 1735--1780, 1997.

\end{thebibliography}

\section{Supplementary Material}

\begin{table}[h]
\scriptsize
	\begin{center}
    \caption{Summary statistics for the benchmark datasets used in our empirical study.
    }

    \label{tab1}
	\begin{sc}
		\begin{tabular}{ccccc}
			\toprule
			Dataset     & Frequency & \# Time Series  & Input Length ($L$)  & Horizon ($K$) \\
			\midrule
			ETTm$_{2}$  & 15 Minute & 7       & 96     & $\{96,192,336,720\}$ \\
	ECL & Hourly    & 321     & 96    & $\{96,192,336,720\}$ \\		
   Exchange    & Daily     & 8       & 96      & $\{96,192,336,720\}$ \\ 
			Traffic    & Hourly    & 862 & 96     & $\{96,192,336,720\}$ \\
			Weather     & 10 Minute & 21      & 96  & $\{96,192,336,720\}$ \\
			ILI         & Weekly    & 7       & 36   & $\{24,36,48,60\}$    \\
			\bottomrule
		\end{tabular}
	\end{sc}
	\end{center}
\end{table}

\begin{table}[h]
\centering
\caption{Multivariate forecasting results for different prediction length $K\in \{96, 192, 336, 720\}$. For the ILI dataset, we set the input length ($L$) to 36, while for the other datasets, we set it to 96. A prediction is considered more precise if it has a lower MSE or MAE value. The metrics are the average of three trials, with the best results highlighted in bold for emphasis.}
\label{tab2}
\resizebox{\textwidth}{!}{%
\begin{tabular}{cccccccccccccccccc}
\toprule
\multicolumn{2}{c}{Models}                                                   & \multicolumn{2}{c}{HGMTS} & \multicolumn{2}{c}{N-BEATS\citep{oreshkin2020N-BEATS}}                     & \multicolumn{2}{c}{Autoformer\citep{wu2021autoformer}}                                          & \multicolumn{2}{c}{Informer\citep{zhou2021informer}}                          & \multicolumn{2}{c}{LongTrans\citep{li2019enhancing}}                         & \multicolumn{2}{c}{Reformer\citep{Kitaev2020Reformer}}                          & \multicolumn{2}{c}{LSTNet\citep{lai2018modeling}}                            & \multicolumn{2}{c}{LSTM\citep{hochreiter1997long}}                         \\ \cline{3-18} 
\multicolumn{2}{c}{Metric}                                                   & MSE                       & MAE                       & MSE                                & MAE                                & MSE                       & MAE                       & MSE                       & MAE                       & MSE                       & MAE                       & MSE                       & MAE                       & MSE                       & MAE                       & MSE                       & MAE                       \\ \hline
\multicolumn{1}{c|}{\multirow{4}{*}{ETTm$_2$}}       & \multicolumn{1}{c|}{96}  & \textbf{0.145}                 & \textbf{0.232}   & 0.184  & 0.263              & 0.255                              & 0.339                              & 0.365                     & 0.453                     & 0.768                     & 0.642                     & 0.658                     & 0.619                     & 3.142                     & 1.365                     & 2.041                     & 1.073                     \\
\multicolumn{1}{c|}{}                             & \multicolumn{1}{c|}{192} & \textbf{0.180}                 & \textbf{0.311}  & 0.273 & 0.337               & 0.281                              & 0.340                              & 0.533                     & 0.563                     & 0.989                     & 0.757                     & 1.078                     & 0.827                     & 3.154                     & 1.369                     & 2.249                     & 1.112                      \\
\multicolumn{1}{c|}{}                             & \multicolumn{1}{c|}{336} & \textbf{0.227}                 & \textbf{0.349}  & 0.309 & 0.355               & 0.339                              & 0.372                              & 1.363                     & 0.887                     & 1.334                     & 0.872                     & 1.549                     & 0.972                     & 3.160                     & 1.369                     & 2.568                     & 1.238                      \\
\multicolumn{1}{c|}{}                             & \multicolumn{1}{c|}{720} & \textbf{0.280}                 & \textbf{0.398}  & 0.411 & 0.425               & 0.422                              & 0.419                              & 3.379                     & 1.388                     & 3.048                     & 1.328                     & 2.631                     & 1.242                     & 3.171                     & 1.368                     & 2.720                     & 1.287       \\ \hline
\multicolumn{1}{c|}{\multirow{4}{*}{ECL}} & \multicolumn{1}{c|}{96}  & \textbf{0.128}            & \textbf{0.226}  & 0.145 & 0.247          & 0.201                              & 0.317                              & 0.274                     & 0.368                     & 0.258                     & 0.357                     & 0.312                     & 0.402                     & 0.680                     & 0.645                     & 0.375                     & 0.437       \\
\multicolumn{1}{c|}{}                             & \multicolumn{1}{c|}{192} & \textbf{0.146}            & \textbf{0.249} & 0.180 & 0.283            & 0.222                              & 0.334                              & 0.296                     & 0.386                     & 0.266                     & 0.368                     & 0.348                     & 0.433                     & 0.725                     & 0.676                     & 0.442                     & 0.473 \\
\multicolumn{1}{c|}{}                             & \multicolumn{1}{c|}{336} & \textbf{0.175}            & \textbf{0.277} & 0.200 & 0.308            & 0.231                              & 0.338                              & 0.300                     & 0.394                     & 0.280                     & 0.380                     & 0.350                     & 0.433                     & 0.828                     & 0.727                     & 0.439                     & 0.473 \\
\multicolumn{1}{c|}{}                             & \multicolumn{1}{c|}{720} & \textbf{0.238}            & \textbf{0.332} & 0.266 & 0.362            & 0.254                              & 0.361                              & 0.373                     & 0.439                     & 0.283                     & 0.376                     & 0.340                     & 0.420                     & 0.957                     & 0.811                     & 0.980                     & 0.814 \\ \hline
\multicolumn{1}{c|}{\multirow{4}{*}{Exchange}}    & \multicolumn{1}{c|}{96}  & \textbf{0.055}            & \textbf{0.172} & 0.098 & 0.206           & 0.197                              & 0.323                              & 0.847                     & 0.752                     & 0.968                     & 0.812                     & 1.065                     & 0.829                     & 1.551                     & 1.058                     & 1.453                     & 1.049 \\
\multicolumn{1}{c|}{}                             & \multicolumn{1}{c|}{192} & \textbf{0.105}            & \textbf{0.242}  & 0.225 & 0.329           & 0.300                              & 0.369                              & 1.204                     & 0.895                     & 1.040                     & 0.851                     & 1.188                     & 0.906                     & 1.477                     & 1.028                     & 1.846                     & 1.179 \\
\multicolumn{1}{c|}{}                             & \multicolumn{1}{c|}{336} & \textbf{0.182}            & \textbf{0.334}  & 0.493 & 0.482          & 0.509                              & 0.524                              & 1.672                     & 1.036                     & 1.659                     & 1.081                     & 1.357                     & 0.976                     & 1.507                     & 1.031                     & 2.136                     & 1.231 \\
\multicolumn{1}{c|}{}                             & \multicolumn{1}{c|}{720} & \textbf{0.560}            & \textbf{0.609} & 1.108 & 0.804           & 1.447                              & 0.941                              & 2.478                     & 1.310                     & 1.941                     & 1.127                     & 1.510                     & 1.016                     & 2.285                     & 1.243                     & 2.984                     & 1.427\\ \hline
\multicolumn{1}{c|}{\multirow{4}{*}{Traffic}}     & \multicolumn{1}{c|}{96}  & \textbf{0.371}                          & \textbf{0.264} & 0.398 & 0.282                         & 0.613                              & 0.388                              & 0.719                     & 0.391                     & 0.684                     & 0.384                     & 0.732                     & 0.423                     & 1.107                     & 0.685                     & 0.843                     & 0.453          \\
\multicolumn{1}{c|}{}                             & \multicolumn{1}{c|}{192} &  \textbf{0.389}               &  \textbf{0.281} & 0.409 & 0.293                        & 0.616                              & 0.382                              & 0.696                     & 0.379                     & 0.685                     & 0.390                     & 0.733                     & 0.420                     & 1.157                     & 0.706                     & 0.847                     & 0.453      \\
\multicolumn{1}{c|}{}                             & \multicolumn{1}{c|}{336} &  \textbf{0.439}                         & \textbf{0.302}  & 0.449 & 0.318                        & 0.622                              & 0.377                              & 0.777                     & 0.420                     & 0.733                     & 0.408                     & 0.742                     & 0.420                     & 1.216                     & 0.730                     & 0.853                     & 0.455 \\
\multicolumn{1}{c|}{}                             & \multicolumn{1}{c|}{720} & \textbf{0.577}                          & \textbf{0.386}   & 0.589 & 0.391                       & 0.660                              & 0.408                              & 0.864                     & 0.472                     & 0.717                     & 0..396                    & 0.755                     & 0.423                     & 1.481                     & 0.805                     & 1.500                     & 0.805   \\ \hline
\multicolumn{1}{c|}{\multirow{4}{*}{Weather}}     & \multicolumn{1}{c|}{96}  & \multicolumn{1}{l}{\textbf{0.146}} & \multicolumn{1}{l}{\textbf{0.185}} &
{0.167} & \multicolumn{1}{l}{0.203} &\multicolumn{1}{l}{0.266} & \multicolumn{1}{c}{0.336} & \multicolumn{1}{l}{0.300} & \multicolumn{1}{l}{0.384} & \multicolumn{1}{l}{0.458} & \multicolumn{1}{l}{0.490} & \multicolumn{1}{l}{0.689} & \multicolumn{1}{l}{0.596} & \multicolumn{1}{l}{0.594} & \multicolumn{1}{l}{0.587} & \multicolumn{1}{l}{0.369} & \multicolumn{1}{l}{0.406} \\
\multicolumn{1}{c|}{}                             & \multicolumn{1}{c|}{192} & \multicolumn{1}{l}{
\textbf{0.207}} & \multicolumn{1}{l}{\textbf{0.236}} & 
{0.229} & \multicolumn{1}{l}{0.261} &
\multicolumn{1}{l}{0.307} & \multicolumn{1}{c}{0.367} & \multicolumn{1}{l}{0.598} & \multicolumn{1}{l}{0.544} & \multicolumn{1}{l}{0.658} & \multicolumn{1}{l}{0.589} & \multicolumn{1}{l}{0.752} & \multicolumn{1}{l}{0.638} & \multicolumn{1}{l}{0.560} & \multicolumn{1}{l}{0.565} & \multicolumn{1}{l}{0.416} & \multicolumn{1}{l}{0.435}  \\
\multicolumn{1}{c|}{}                             & \multicolumn{1}{c|}{336} & \multicolumn{1}{l}{\textbf{0.268}} & \multicolumn{1}{l}{\textbf{0.291}} & 
{0.287} & \multicolumn{1}{l}{0.304} &
\multicolumn{1}{l}{0.359} & \multicolumn{1}{c}{0.395} & \multicolumn{1}{l}{0.578} & \multicolumn{1}{l}{0.523} & \multicolumn{1}{l}{0.797} & \multicolumn{1}{l}{0.652} & \multicolumn{1}{l}{0.639} & \multicolumn{1}{l}{0.596} & \multicolumn{1}{l}{0.597} & \multicolumn{1}{l}{0.587} & \multicolumn{1}{l}{0.455} & \multicolumn{1}{l}{0.454}  \\
\multicolumn{1}{c|}{}                             & \multicolumn{1}{c|}{720} & \multicolumn{1}{l}{\textbf{0.348}} & \multicolumn{1}{l}{\textbf{0.351}} & 
{0.368} & \multicolumn{1}{l}{0.359} &
\multicolumn{1}{l}{0.419} & \multicolumn{1}{c}{0.428} & \multicolumn{1}{l}{1.059} & \multicolumn{1}{l}{0.741} & \multicolumn{1}{l}{0.869} & \multicolumn{1}{l}{0.675} & \multicolumn{1}{l}{1.130} & \multicolumn{1}{l}{0.792} & \multicolumn{1}{l}{0.618} & \multicolumn{1}{l}{0.599} & \multicolumn{1}{l}{0.535} & \multicolumn{1}{l}{0.520}  \\ \hline
\multicolumn{1}{c|}{\multirow{4}{*}{ILI}}     & \multicolumn{1}{c|}{24}  & \textbf{1.827}                          & \textbf{0.839}                          & 1.879 & 0.886 & 3.483                              & 1.287                              & 5.764                     & 1.677                     & 4.480                     & 1.444                     & 4.400                     & 1.382                     & 6.026                     & 1.770                     & 5.914                     & 1.734       \\
\multicolumn{1}{c|}{}                             & \multicolumn{1}{c|}{36} & \textbf{2.034}                          & \textbf{0.903} & 2.210 & 1.018                         & 3.103                              & 1.148                              & 4.755                     & 1.467                     & 4.799                     & 1.467                     & 4.783                     & 1.448                     & 5.340                     & 1.668                     & 6.631                     & 1.845          \\
\multicolumn{1}{c|}{}                             & \multicolumn{1}{c|}{48} & \textbf{2.102}                          & \textbf{0.915} & 2.440 & 1.088                         & 2.669                              & 1.085                              & 4.763                     & 1.469                     & 4.800                     & 1.468                     & 4.832                     & 1.465                     & 6.080                     & 1.787                     & 6.736                     & 1.857          \\
\multicolumn{1}{c|}{}                             & \multicolumn{1}{c|}{60} & \textbf{2.118}                          & \textbf{0.956} & 2.547 & 1.057                         & 2.770                              & 1.125                              & 5.264                     & 1.564                     & 5.278                     & 1.560                    & 4.882                     & 1.483                     & 5.548                     & 1.720                     & 6.870                     & 1.879          \\ 
\bottomrule
\end{tabular}
}
\end{table}

\begin{table}[h]
\centering
\caption{The HGMTS performance evaluated under various selections of the graph sparsity hyper-parameter $\gamma$. The forecasting setup remains consistent with what is presented in Table \ref{tab2}.}
\label{tab3}
\resizebox{\textwidth}{!}{%
\begin{tabular}{cccccccccccccc}
\toprule
\multicolumn{2}{c}{Sparsity ($\gamma$)}                                                   & \multicolumn{2}{c}{0.2} & \multicolumn{2}{c}{0.3}                     & \multicolumn{2}{c}{0.4}                                          & \multicolumn{2}{c}{0.5}                          & \multicolumn{2}{c}{0.6}                         & \multicolumn{2}{c}{0.7}                      \\ \cline{3-14} 
\multicolumn{2}{c}{Metric}                                                   & MSE                       & MAE                       & MSE                                & MAE                                & MSE                       & MAE                       & MSE                       & MAE                       & MSE                       & MAE                       & MSE                       & MAE                                             \\ \hline
\multicolumn{1}{c|}{\multirow{4}{*}{ETTm$_2$}}       & \multicolumn{1}{c|}{96}  & 0.151                 & 0.240   & 0.149  & 0.238 & 0.146 & 0.234 & \textbf{0.145}  & \textbf{0.232}  & --   & -- & 0.146  & 0.235  \\
\multicolumn{1}{c|}{}                             & \multicolumn{1}{c|}{192} & 0.187                 & 0.324  & 0.183 & 0.315             & 0.181                 & 0.313                        & \textbf{0.180}                 & \textbf{0.311}        & --                     & --                     & 1.182                     & 0.314                              \\
\multicolumn{1}{c|}{}                             & \multicolumn{1}{c|}{336} & 0.239                 & 0.364  & 0.234 & 0.358             & 0.230                 & 0.353                        & \textbf{0.227}                 & \textbf{0.349}        & --                     & --                     & 0.229                     & 0.352                              \\
\multicolumn{1}{c|}{}                             & \multicolumn{1}{c|}{720} & 0.292                 & 0.418  & 0.286 & 0.407             & 0.283                 & 0.402                        & \textbf{0.280}                 & \textbf{0.398}        & --                     & --                     & 0.282                     & 0.401               \\ \hline
\multicolumn{1}{c|}{\multirow{4}{*}{ECL}} & \multicolumn{1}{c|}{96}  & \textbf{0.128}            & \textbf{0.226}  & 0.130 & 0.229          & 0.133                              & 0.234                              & 0.138                     & 0.241                     & 0.145                     & 0.249                     & 0.156                     & 0.263               \\
\multicolumn{1}{c|}{}                             & \multicolumn{1}{c|}{192} & \textbf{0.146}            & \textbf{0.249} & 0.149 & 0.253            & 0.152                              & 0.257                              & 0.158                     & 0.266                     & 0.164                     & 0.274                     & 0.169                     & 0.280         \\
\multicolumn{1}{c|}{}                             & \multicolumn{1}{c|}{336} & \textbf{0.175}            & \textbf{0.277} & 0.177 & 0.280            & 0.181                              & 0.285                              & 0.189                     & 0.294                     & 0.193                     & 0.301                     & 0.198                     & 0.307         \\
\multicolumn{1}{c|}{}                             & \multicolumn{1}{c|}{720} & \textbf{0.238}            & \textbf{0.332} & 0.240 & 0.335            & 0.243                              & 0.338                              & 0.247                     & 0.345                     & 0.252                     & 0.351                     & 0.256                     & 0.359         \\ \hline
\multicolumn{1}{c|}{\multirow{4}{*}{Exchange}}    & \multicolumn{1}{c|}{96}  & --     & -- & 0.056 & 0.173         &  \textbf{0.055}            & \textbf{0.172}                      & 0.057                     & 0.174                     & 0.058                     & 0.176                     & 0.061                     & 0.180         \\
\multicolumn{1}{c|}{}                             & \multicolumn{1}{c|}{192} & --       & --  & 0.107 & 0.244          & \textbf{0.105}            & \textbf{0.242}                            & 0.106                     & 0.244                     & 0.107                     & 0.246                     & 0.109                     & 0.249         \\
\multicolumn{1}{c|}{}                             & \multicolumn{1}{c|}{336} & --         & --  & 0.184 & 0.336          & \textbf{0.182}            & \textbf{0.334}                      & 0.183                     & 0.336                     & 0.184                     & 0.338                     & 0.187                     & 0.341         \\
\multicolumn{1}{c|}{}                             & \multicolumn{1}{c|}{720} & --     & -- & 0.563 & 0.613           & \textbf{0.560}            & \textbf{0.609}                      & 0.562                     & 0.611                     & 0.564                     & 0.613                     & 0.567                     & 0.617        \\ \hline
\multicolumn{1}{c|}{\multirow{4}{*}{Traffic}}     & \multicolumn{1}{c|}{96}  & \textbf{0.371}                          & \textbf{0.264} & 0.374 & 0.268                         & 0.377                              & 0.272                              & 0.381                     & 0.276                     & 0.386                     & 0.280                     & 0.391                     & 0.287          \\
\multicolumn{1}{c|}{}                             & \multicolumn{1}{c|}{192} &  \textbf{0.389}               &  \textbf{0.281} & 0.394 & 0.286                        & 0.398                              & 0.291                              & 0.405                     & 0.299                     & 0.412                     & 0.308                     & 0.423                     & 0.316                    \\
\multicolumn{1}{c|}{}                             & \multicolumn{1}{c|}{336} &  \textbf{0.439}                         & \textbf{0.302}  & 0.445 & 0.309                        & 0.451                              & 0.316                              & 0.460                     & 0.327                     & 0.463                     & 0.332                     & 0.466                     & 0.338    \\
\multicolumn{1}{c|}{}                             & \multicolumn{1}{c|}{720} & \textbf{0.577}                          & \textbf{0.386}   & 0.581 & 0.392                       & 0.584                              & 0.395                              & 0.589                     & 0.401                     & 0.594                     & 0.407                    & 0.598                     & 0.414   \\ \hline
\multicolumn{1}{c|}{\multirow{4}{*}{Weather}}     & \multicolumn{1}{c|}{96}  & \multicolumn{1}{l}{0.147} & \multicolumn{1}{l}{0.186} & \multicolumn{1}{l}{\textbf{0.146}} & \multicolumn{1}{l}{\textbf{0.185}} &\multicolumn{1}{l}{0.147} & \multicolumn{1}{c}{0.187} & \multicolumn{1}{l}{0.149} & \multicolumn{1}{l}{0.190} & \multicolumn{1}{l}{0.150} & \multicolumn{1}{l}{0.191} & \multicolumn{1}{l}{0.152} & \multicolumn{1}{l}{0.193}  \\
\multicolumn{1}{c|}{}                             & \multicolumn{1}{c|}{192} & \multicolumn{1}{l}{
0.208} & \multicolumn{1}{l}{0.238} & 
\multicolumn{1}{l}{
\textbf{0.207}} & \multicolumn{1}{l}{\textbf{0.236}} &
\multicolumn{1}{l}{0.209} & \multicolumn{1}{c}{0.238} & \multicolumn{1}{l}{0.212} & \multicolumn{1}{l}{0.240} & \multicolumn{1}{l}{0.215} & \multicolumn{1}{l}{0.244} & \multicolumn{1}{l}{0.218} & \multicolumn{1}{l}{0.248}   \\
\multicolumn{1}{c|}{}                             & \multicolumn{1}{c|}{336} & \multicolumn{1}{l}{0.270} & \multicolumn{1}{l}{0.293}  
& \multicolumn{1}{l}{\textbf{0.268}} & \multicolumn{1}{l}{\textbf{0.291}} &
\multicolumn{1}{l}{0.269} & \multicolumn{1}{c}{0.292} & \multicolumn{1}{l}{0.271} & \multicolumn{1}{l}{0.294} & \multicolumn{1}{l}{0.274} & \multicolumn{1}{l}{0.298} & \multicolumn{1}{l}{0.277} & \multicolumn{1}{l}{0.301}  \\
\multicolumn{1}{c|}{}                             & \multicolumn{1}{c|}{720} & \multicolumn{1}{l}{0.350} & \multicolumn{1}{l}{0.354} & 
\multicolumn{1}{l}{\textbf{0.348}} & \multicolumn{1}{l}{\textbf{0.351}} & 
\multicolumn{1}{l}{0.349} & \multicolumn{1}{c}{0.353} & \multicolumn{1}{l}{0.352} & \multicolumn{1}{l}{0.356} & \multicolumn{1}{l}{0.354} & \multicolumn{1}{l}{0.359} & \multicolumn{1}{l}{0.357} & \multicolumn{1}{l}{0.364}   \\ \hline
\multicolumn{1}{c|}{\multirow{4}{*}{ILI}}     & \multicolumn{1}{c|}{24}  & 1.832                          & 0.845              & 1.830 & 0.842 & 1.828                   & 0.840                              & \textbf{1.827}                          & \textbf{0.839}                     & -- & --                     & 1.829          & 0.842     \\
\multicolumn{1}{c|}{}                             & \multicolumn{1}{c|}{36} & 2.041                          & 0.911 & 2.037 & 0.906                         & 2.036            & 0.905                              & \textbf{2.034}                          & \textbf{0.903}                     & -- & --                     & 2.035      & 0.905       \\
\multicolumn{1}{c|}{}                             & \multicolumn{1}{c|}{48} & 2.113                          & 0.926 & 2.108 & 0.922                         & 2.105            & 0.918                              & \textbf{2.102}                          & \textbf{0.915}                     & -- & --                     & 2.104      & 0.918         \\
\multicolumn{1}{c|}{}                             & \multicolumn{1}{c|}{60} & 2.123                          & 0.964 & 2.122 & 0.962                         & 2.120          & 0.959                              & \textbf{2.118}                          & \textbf{0.956}                     & -- & --                    & 2.119      & 0.959       \\ 
\bottomrule
\end{tabular}
}
\end{table}

\begin{table}[h] 
\scriptsize
    \begin{center}
    \caption{Empirical evaluation of long sequence time series forecasts for HGMTS. MAE and MSE are averaged over three runs and five datasets, with the best result highlighted in bold and the second best in blue.}
    \label{tab4}
	\begin{tabular}{ll | cccccc} \toprule
    						&     & HGMTS$_{1}$                   & HGMTS$_{2}$             & HGMTS$_{3}$              & HGMTS$_{4}$ & HGMTS$_{5}$ &
                            HGMTS$_{6}$\\ \midrule
\parbox[t]{1mm}{\multirow{4}{*}{\rotatebox[origin=c]{90}{A. MSE}}}
							& 96  & \textbf{0.168}  & 0.171                   & 0.170           & 0.195       & 0.183 & \textcolor{blue}{0.169}    \\
							& 192 & \textbf{0.205}           & 0.209                   & 0.208  & 0.261       & 0.232 & \textcolor{blue}{0.206}    \\
							& 336 & \textbf{0.258}           & \textcolor{blue}{0.263} & 0.271                    & 0.344       & 0.309 & 0.264    \\
							& 720 & \textbf{0.401}           & \textcolor{blue}{0.412} & 0.428                    & 0.545       & 0.496 & 0.414    \\ \midrule
\parbox[t]{1mm}{\multirow{4}{*}{\rotatebox[origin=c]{90}{A. MAE}}}
							& 96  & \textbf{0.214}  & 0.219                   & 0.218           & 0.237       & 0.229 & \textcolor{blue}{0.216}  \\
							& 192 & \textbf{0.264}           & 0.268                   & 0.266  & 0.296       & 0.286 & \textcolor{blue}{0.265}  \\
							& 336 & \textbf{0.311}           & \textcolor{blue}{0.316} & 0.328                    & 0.349       & 0.337 & 0.318  \\
							& 720 & \textbf{0.415}           & \textcolor{blue}{0.418} & 0.421                    & 0.464       & 0.435 & 0.420  \\ \bottomrule
	\end{tabular}
	\end{center}
\end{table}

\end{document}